\documentclass[10pt,twocolumn,letterpaper]{article}

\usepackage{iccv}
\usepackage{times}
\usepackage{epsfig}
\usepackage{graphicx}
\usepackage{amsmath}
\usepackage{amssymb}
\usepackage{amsmath}
\usepackage{amsfonts}

\usepackage{multirow}
\usepackage{multicol}
\usepackage{booktabs}
\usepackage{array}
\usepackage{caption}
\usepackage{subcaption}
\usepackage{arydshln}

\usepackage{balance}

\usepackage[accsupp]{axessibility}

\usepackage[pagebackref=true,breaklinks=true,letterpaper=true,colorlinks,bookmarks=false]{hyperref}

\iccvfinalcopy 

\ificcvfinal\fi

\begin{document}

\title{DOLG: Single-Stage Image Retrieval with Deep Orthogonal Fusion of Local and Global Features}

\title{DOLG: Single-Stage Image Retrieval with \textit{D}eep \textit{O}rthogonal Fusion of\\ \textit{L}ocal and \textit{G}lobal Features}

\author{Min Yang$^{*}$, Dongliang He$^{*,\dag}$, Miao Fan, Baorong Shi,\\ Xuetong Xue, Fu Li, Errui Ding, Jizhou Huang$^{\dag}$\\
Baidu Inc., China\\
{\tt\small \{yangmin09, hedongliang01, fanmiao, shibaorong\}}@baidu.com,\\ {\tt\small\{xuexuetong, lifu, dingerrui, huangjizhou01\}}@baidu.com
}

\maketitle

\renewcommand{\thefootnote}{\fnsymbol{footnote}}
\footnotetext[1]{~Equal contribution. $^\dag$~Corresponding authors.}
\renewcommand{\thefootnote}{\arabic{footnote}}

\ificcvfinal\thispagestyle{empty}\fi

\begin{abstract}
Image Retrieval is a fundamental task of obtaining images similar to the query one  from a database. A common image retrieval practice is to firstly retrieve candidate images via similarity search using global image features and then re-rank the candidates by leveraging their local features. Previous learning-based studies mainly focus on either global or local image representation learning to tackle the retrieval task. In this paper, we abandon the two-stage paradigm and seek to design an effective single-stage solution by integrating local and global information inside images into compact image representations. Specifically, we propose a Deep Orthogonal Local and Global (DOLG) information fusion framework for end-to-end image retrieval. It attentively extracts representative local information with multi-atrous convolutions and self-attention at first. Components orthogonal to the global image representation are then extracted from the local information. At last, the orthogonal components are concatenated with the global representation as a complementary, and then aggregation is performed to generate the final representation. The whole framework is end-to-end differentiable and can be trained with image-level labels. Extensive experimental results validate the effectiveness of our solution and show that our model achieves state-of-the-art image retrieval performances on Revisited Oxford and Paris datasets. \footnote{Codes: \href{https://github.com/feymanpriv/DOLG-paddle}{PaddlePaddle Implementation}.}
\end{abstract}

\section{Introduction}
\begin{figure}[t]
  \centering
  \includegraphics[width=1.\columnwidth]{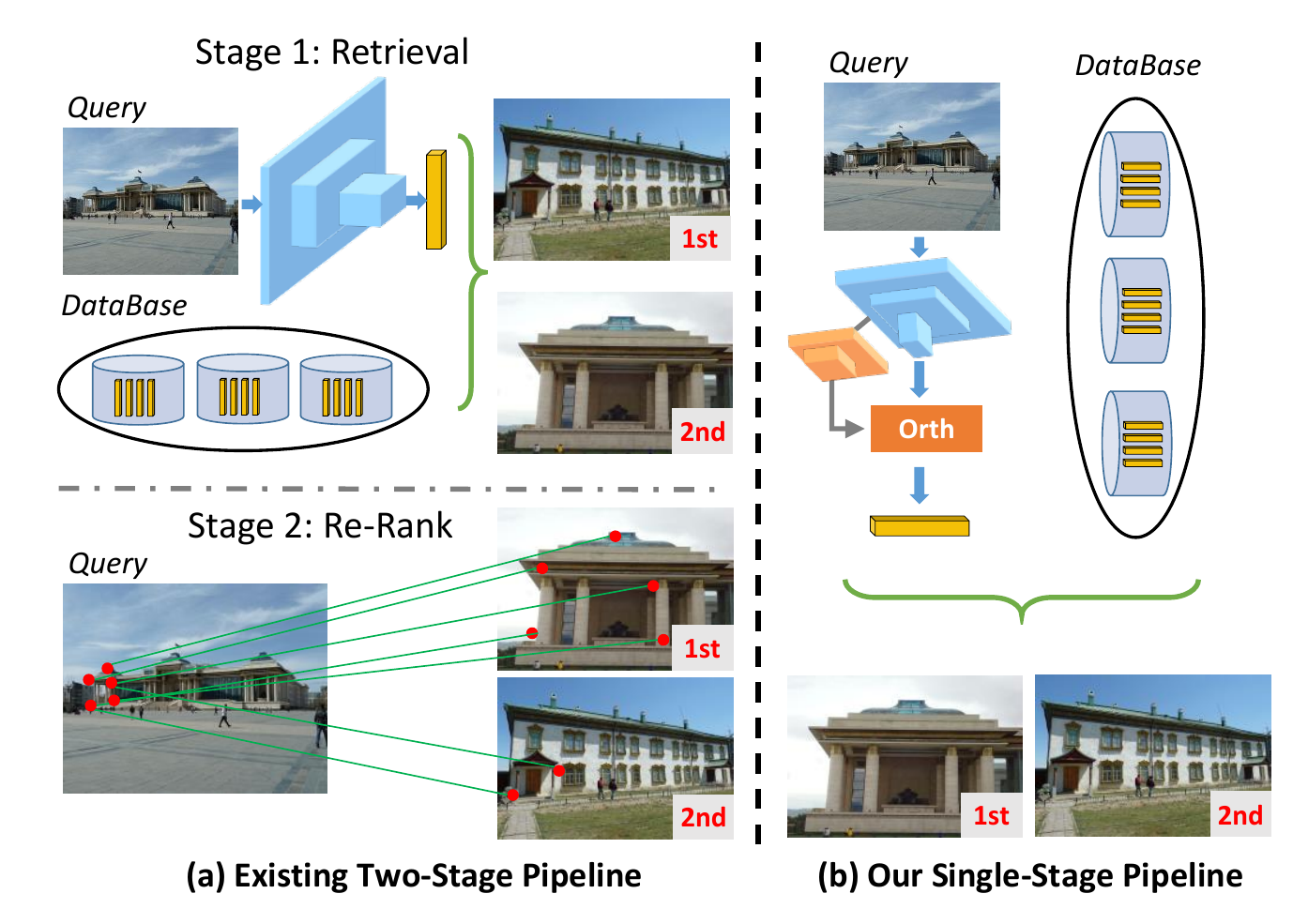}
   \caption{Illustration of current two-stage and our single-stage image retrieval. Previous methods (a) firstly obtain candidates similar to the query from the database via global deep representation, and then local descriptors are extracted for leveraged re-ranking. Our method (b) aggregates global and local features via an orthogonal fusion to generate the final compact descriptor, and then single-shot similarity search is performed.}
\label{illustration}
\end{figure}
Image retrieval is an important task in computer vision, and its main purpose is to find out the images from a large-scale database that are similar to a query one. It is extensively studied by designing various handcrafted features \cite{lowe2004distinctive,bay2008speeded,zheng2009tour}. Owing to the development of deep learning technologies, great progress has been achieved recently \cite{arandjelovic2016netvlad,noh2017large,sarlin2019coarse,cao2020unifying}. Representations (also named as descriptors) of images, which are used to encode image contents and measure their similarities, play a central role in this task. In the literature of learning-based solutions, two types of image representations are widely explored. One is global feature \cite{babenko2014neural, babenko2015aggregating, tolias2015particular, arandjelovic2016netvlad} which serves as high-level semantic image signature and the other one is local feature \cite{balntas2016learning,revaud2019r2d2,noh2017large,he2018local} which can comprise discriminative geometry information about specific image regions. Generally, the global feature can be learned to be invariant to viewpoint and illumination, while local features are more sensitive to local geometry and textures. Therefore, previous state-of-the-art solutions \cite{dsm,noh2017large,cao2020unifying} always work in a two-stage paradigm. As shown in Figure \ref{illustration}(a), candidates are retrieved via global feature with high recall, and then re-ranking is performed with local features to further improve precision. 

In this paper, we also concentrate on the field of image retrieval with deep networks. Though state-of-the-art performance has been achieved by previous two-stage solutions, they need to rank images twice, and the second re-ranking stage is conducted using the expensive RANSAC \cite{ransac} or AMSK \cite{tolias2016image} for spatial verification with local features. More importantly, errors exist inevitably in both stages. 
Two-stage solutions would suffer from error accumulation which can be a bottleneck for further performance improvement. 
To alleviate these problems, we abandon the two-stage framework and attempt to find an effective unified single-stage image retrieval solution, which is shown in Figure \ref{illustration}(b). 
Previous wisdom has implied that global features and local features are two complementary and essential elements for image retrieval. Intuitively, integrating local features and global features into a compact descriptor can achieve our goal. A satisfying local and global fusion scheme can take advantage of both types of features to mutually boost each other for single-stage retrieval. Besides, error accumulation can be avoided.
Therefore, we technically answer how to design an effective global and local fusion mechanism for end-to-end single-stage image retrieval.

Specifically, we proposed a \textbf{D}eep \textbf{O}rthogonal \textbf{L}ocal and \textbf{G}lobal  feature fusion model (\textit{DOLG}). It consists of a local and a global branch for learning two types of features jointly and an orthogonal fusion module to combine them. In detail, the local components orthogonal to the global feature are decomposed from the local features. Subsequently, the orthogonal components are concatenated with the global feature as a complementary part. Finally, it is aggregated into a compact descriptor. 
With our orthogonal fusion, the most critical local information can be extracted and redundant components to the global information are eliminated, such that local and global components can be mutually reinforced to produce final representative descriptor with objective-oriented training. 
To enhance local feature learning, inspired by lessons from prior research, the local branch is equipped with multi-atrous convolutions \cite{atrous} and self-attention \cite{noh2017large} mechanisms to attentively extract representative local features. We think alike FP-Net \cite{fpnet} in terms of orthogonal feature space learning, but DOLG aims at complementary fusion of features in orthogonal spaces. 
Extensive experiments on Revisited Oxford and Pairs \cite{radenovic2018revisiting} show the effectiveness of our framework. DOLG also achieves state-of-the-art performance on both datasets. 
To summarize, our main contributions are as follows:
\begin{itemize}
    \item We propose to retrieve images in a single-stage paradigm with a novel orthogonal global and local feature fusion framework, which can generate a compact representative image descriptor and is end-to-end learnable. 
    \item In order to attentively extract discriminative local features, a module with multi-atrous convolution layers followed by a self-attention module is designed for improving our local branch.
    \item Extensive experiments are conducted and comprehensive analysis is provided to validate the effectiveness of our solution. Our single-stage method significantly outperforms previous two-stage state-of-the-art ones.
\end{itemize}

\section{Related Work}

\subsection{Local feature}

Prior to deep learning, SIFT \cite{lowe2004distinctive} and SURF \cite{bay2008speeded} are two well-known hand-engineered local features. 
Usually such local features are combined with KD trees \cite{beis1997shape}, vocabulary trees \cite{nister2006scalable}  or encoded by aggregation methods such as \cite{bagofwords,vlad} for (approximate) nearest neighbor search. 
Spatial verification via matching local features with RANSAC \cite{ransac} to re-rank candidate retrieval results \cite{avrithis2014hough,philbin2007object} are also shown to significantly improve precision.
Recently, driven by the development of deep learning, remarkable progresses have been made in learning local features from images such as  
\cite{yi2016lift,dusmanu2019d2,detone2018superpoint,balntas2016learning,revaud2019r2d2,noh2017large,he2018local}. Comprehensive reviews of deep local feature learning can be found in \cite{zhou2017recent,chen2021deep}. 
Among these methods, the state-of-the-art local feature learning framework DELF \cite{noh2017large}, which proposes an attentive local feature descriptor for large-scale image retrieval, is closely related to our work. One of the design choices of our local branch, namely attentive feature extraction, is inspired by its merit. However, DELF uses only a single-scale feature map and ignores various object scales inside natural images. Our local branch is designed to simulate the image pyramid trick used in SIFT \cite{lowe2004distinctive} by multi-atrous convolution layers \cite{atrous}. 

\begin{figure*}[t]
\centering
\includegraphics[width=0.85\textwidth]{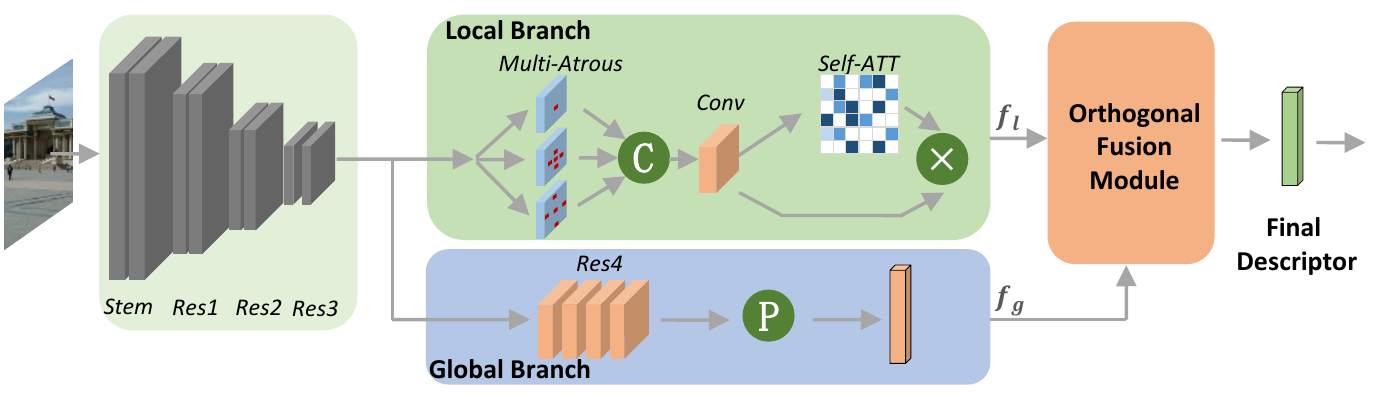}
\caption{Block diagram of our deep orthogonal local and global (DOLG) information fusion framework. Taking ResNet \cite{resnet} for illustraction, we build a local branch and a global branch after Res3. The local branch uses multi-atrous layers to simulate spatial pyramid to take into consideration of scale variations among images. Self-attention is leveraged for importance modeling following lessons of existing works \cite{noh2017large,cao2020unifying}. The global branch generates a descriptor, which is fed into an orthogonal fusion module together with the local features for integrating both types of features into a final compact descriptor. ``P'', ``C'' and ``X'' denote pooling, concatenation and element-wise multiplication, respectively.}
\label{fig:arch}
\end{figure*}

\subsection{Global feature}
Conventional solutions obtain global feature by aggregating local features by BoW \cite{sivic2003video, radenovic2016cnn}, Fisher vectors \cite{jegou2011aggregating} or VLAD \cite{jegou2010aggregating}. Later, aggregated selective match kernels (ASMK) \cite{tolias2016image} attempts to unify aggregation-based techniques with matching-based approaches such as Hamming Embedding \cite{jegou2008hamming}.
In deep learning era, global feature is obtained by such differentiable aggregation operations as sum-pooling \cite{tolias2020learning} and GeM pooling \cite{radenovic2018fine}. To train deep CNN models, ranking based triplet \cite{triplet}, quadruplet \cite{quadruplet}, angular \cite{angular} and listwise \cite{revaud2019learning} losses or classification based  losses \cite{wang2018cosface, deng2019arcface} are proposed. With these innovations, nowadays, most high performing global features are obtained with deep CNNs for image retrieval\cite{babenko2014neural, babenko2015aggregating, tolias2015particular, arandjelovic2016netvlad, gordo2017end, radenovic2018fine, revaud2019learning,noh2017large,ng2020solar,cao2020unifying}. In our work, we leverage lessons from previous studies to use ArcFace loss \cite{deng2019arcface} in the training phase and to explore different pooling schemes for performance improvement. Our model also generates a compact descriptor, meanwhile, it explicitly considers fusing local and global features in an orthogonal way.

\subsection{Joint local and global CNN features}
It is natural to consider local and global features jointly, because feature maps from an image representation model can be interpreted as local visual words \cite{dsm,taira2018inloc}. Joint learning local matching and global representation may be beneficial for both sides. Therefore, distilling pre-trained local feature \cite{detone2018superpoint} and global feature \cite{arandjelovic2016netvlad} into a compact descriptor is proposed in \cite{sarlin2019coarse}. DELG \cite{cao2020unifying} takes a step further and proposes to jointly train local and global features in an end-to-end manner. However, DELG still works in a two-stage fashion. Our work is essentially different from \cite{noh2017large,cao2020unifying} and we propose orthogonal global and local fusion in order to perform accurate single-stage image retrieval.

\section{Methodology}

\subsection{Overview}
Our DOLG framework is depicted in Figure \ref{fig:arch}. Following \cite{noh2017large,cao2020unifying}, it is built upon state-of-the-art image recognition model ResNet \cite{resnet}. The global branch is kept the same as the original ResNet except that 1) the global averaging pooling is replaced by the GeM pooling \cite{radenovic2018fine}; 2) a FC layer is used to reduce feature dimension when generating the global representation $f_g\in R^{C\times1}$. 
Specifically, let us denote the output feature map of Res4 as $f_4\in R^{C_4\times h \times w}$, then the GeM pooling can be formalized as
\begin{equation}
    f_{g,c} =  \left ( \frac{1}{hw}\sum_{(i,j)}f_{4, (c,i,j)}^{p} \right )^{1/p}_{c=1,2,...,C_4},
\end{equation}
where $p>0$ is a hyper-parameter and $p>1$ pushes the output to focus more on salient feature points. In this paper, we follow the setting of DELG \cite{cao2020unifying} and empirically set it to be 3.0.
To jointly extract local descriptors, a local branch is appended after the Res3 block of ResNet. Our local branch consists of multiple atrous convolution layers \cite{atrous} and a self-attention module. Then, a novel orthogonal fusion module is designed for aggregating $f_g$ and the local feature tensor $f_l\in R^{C\times H \times W}$ obtained by the local branch. After orthogonal fusion, a final compact descriptor, where local and global information is well integrated, is generated.

\subsection{Local Branch}
\begin{figure*}[t]
\centering
\includegraphics[width=0.75\textwidth]{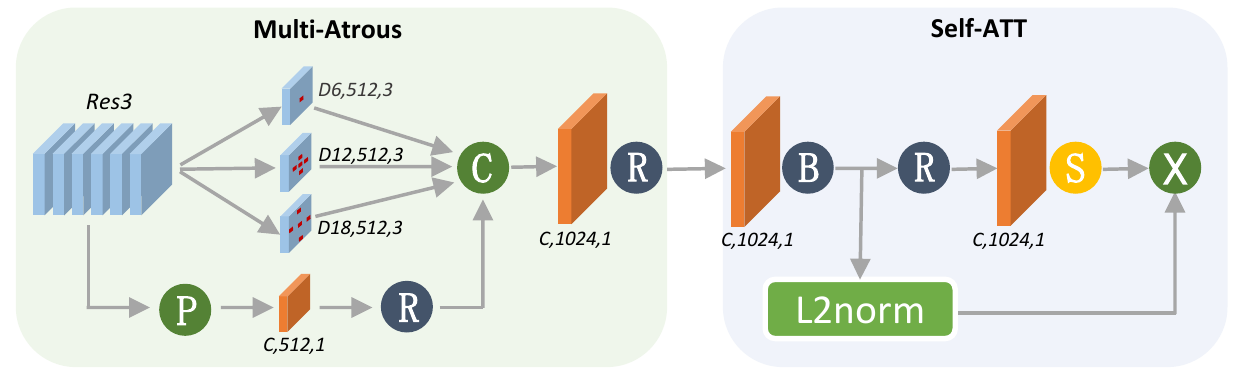}
\caption{Configurations of our local branch. ``D$s,c,k$'' denotes dilated convolution with rate $s$, output channel number $c$ and kernel size $k$. ``C,$c,k$'' means vanilla convolution. ``R'', ``B'' and ``S'' denote ReLU, BN and Softplus, respectively.}
\label{ma}
\end{figure*}
The two major building blocks of our local branch are the multi-atrous convolution layers and the self-attention module. The former building block is to simulate feature pyramid which can handle scale variations among different image instances, and the latter building block is leveraged to performance importance modeling. 
The detailed network configurations of this branch is shown in Figure \ref{ma}. The multi-atrous module contains three dilated convolution layers to obtain feature maps with different spatial receptive field and a global average pooling branch. 
These features are concatenated and then processed by a $1\times1$ convolution layer. The output feature map is then delivered to the self-attention module for further modeling the importance of each local feature point. Specifically, its input is firstly processed using a $1\times1$ conv-bn module, then the subsequent feature is normalized and modulated by an attention map generated via a  $1\times1$ convolution layer followed by the SoftPlus operation.

\begin{figure}[t]
\centering
\begin{subfigure}[b]{0.48\columnwidth}
\centering
\includegraphics[width=\columnwidth]{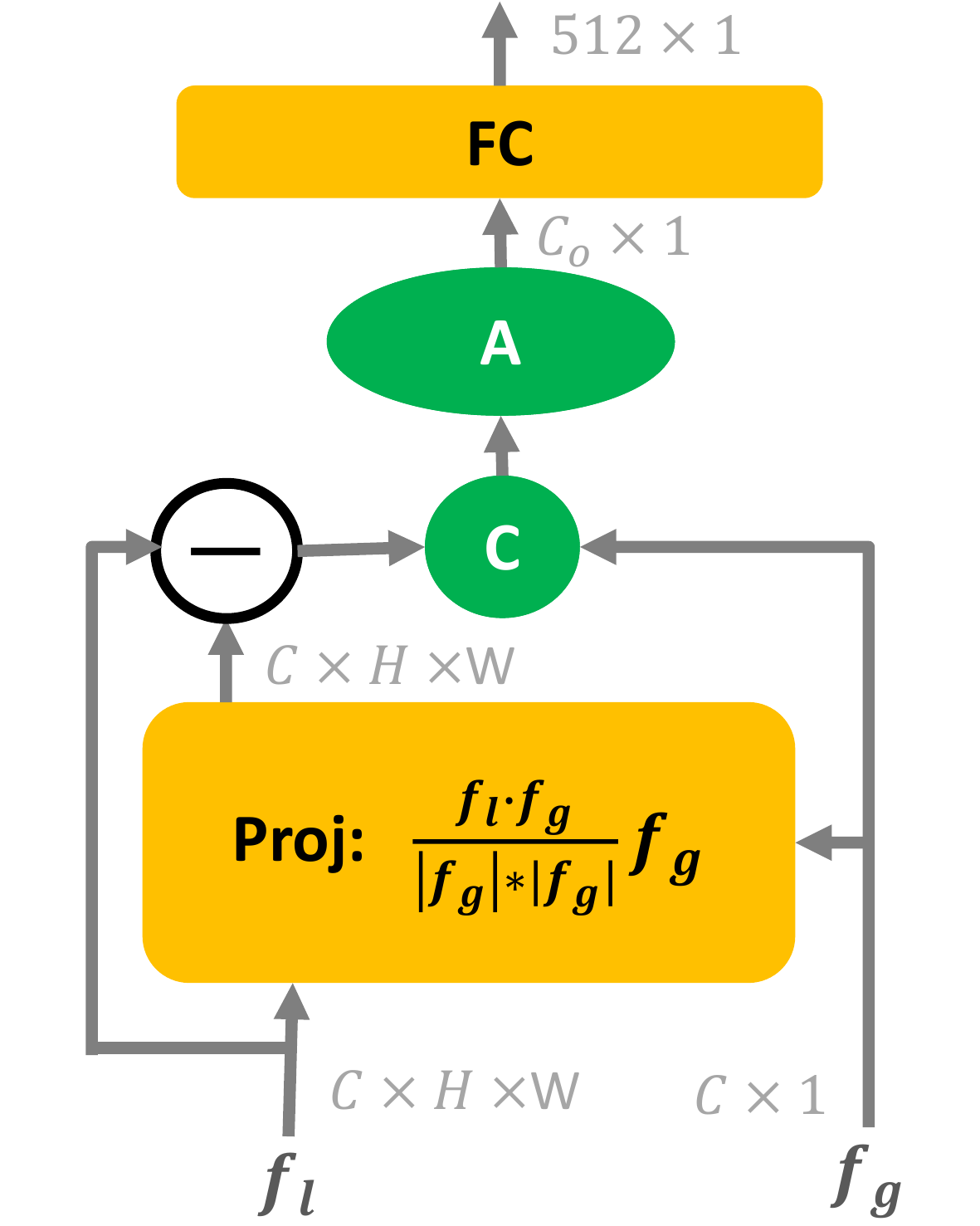}
\caption{Framework of our proposed orthogonal fusion module. ``A'' denotes aggregation.}
\label{fig:ortho1}
\end{subfigure}
\hfill
\begin{subfigure}[b]{0.43\columnwidth}
\centering
\includegraphics[width=\columnwidth]{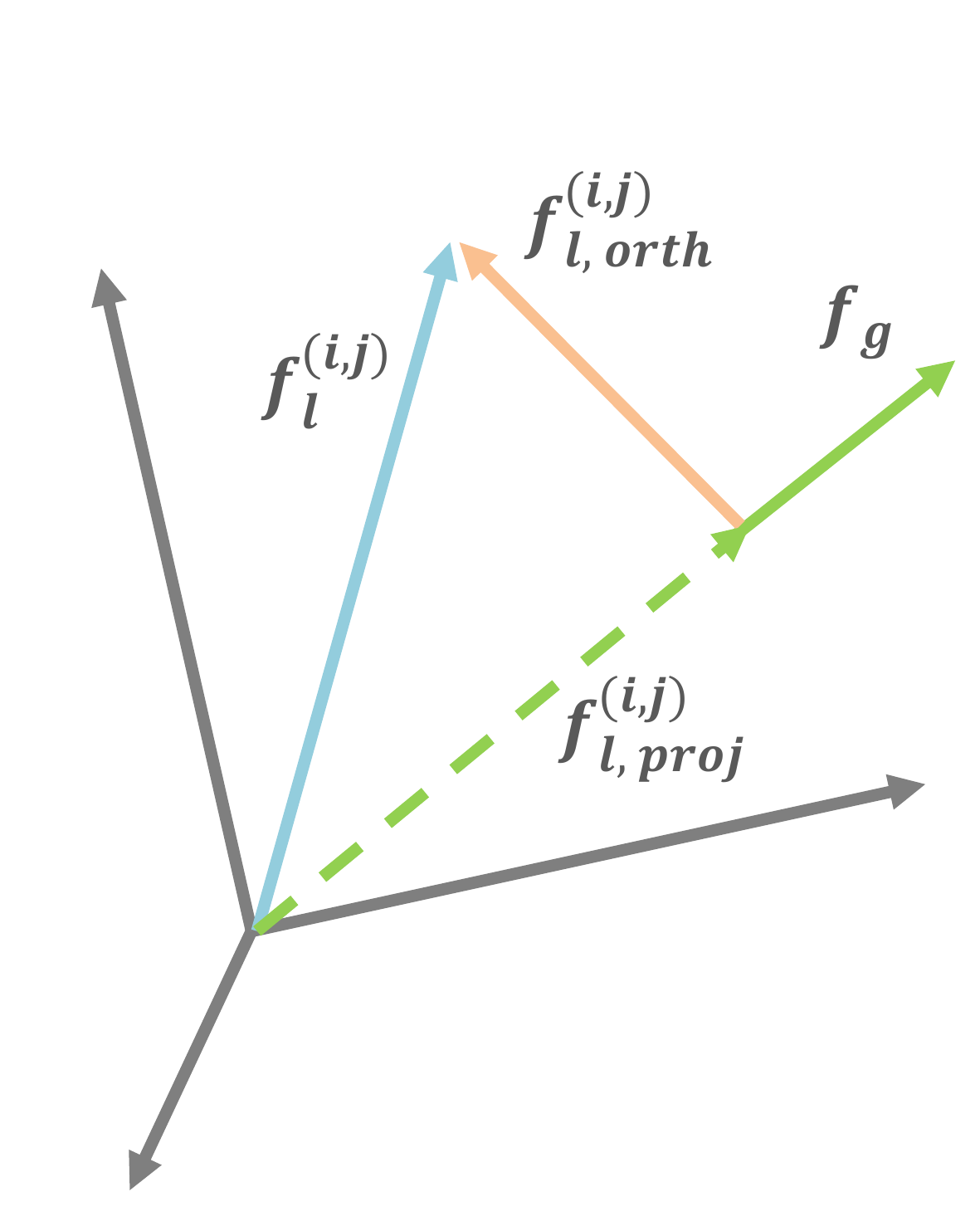}
\caption{Demonstration of a local feature projected on the global feature and the component orthogonal to the global feature.}
\label{fig:ortho2}
\end{subfigure}
\end{figure}

\subsection{Orthogonal Fusion Module}
The working flow of our orthogonal fusion module is shown in Figure \ref{fig:ortho1}. It takes $f_l$ and $f_g$ as inputs and then calculates the projection $f_{l,proj}^{(i,j)}$ of each local feature point $f_l^{(i,j)}$ onto the global feature $f_g$. Mathematically,
the projection can be formulated as:
\begin{equation}
    f_{l,proj}^{(i,j)} = \frac{f_l^{(i,j)} \cdot f_g}{|f_g|^2}f_g,
\end{equation}
where $f_l^{(i,j)}\cdot f_g$ is dot product operation and $|f_g|^2$ is the $L_2$ norm of $f_g$:
\begin{equation}
    f_l^{(i,j)}\cdot f_g = \Sigma_{c=1}^C{f_{l,c}^{(i,j)}f_{g,c}}
\end{equation}
\begin{equation}
    |f_g|^2 = \Sigma_{c=1}^C{(f_{g,c})^2}.
\end{equation}
As demonstrated in Figure \ref{fig:ortho2}, the orthogonal component is the difference between the local feature and its projection vector, therefore, we can obtain the component orthogonal to $f_g$ by:
\begin{equation}
    f_{l,orth}^{i,j} = f_l^{(i,j)} - f_{l,proj}^{(i,j)}.
\end{equation}

In this way, a $C\times H \times W$ tensor where each point is orthogonal to $f_g$ can be extracted. Afterwards, we append to each point of this tensor with the $C\times1$ vector $f_g$ and then the new tensor is aggregated to be a $C_o\times1$ vector. Finally, a fully connected layer is used to produce a $512\times1$ descriptor. Typically, $C$ equals 1024 in ResNet \cite{resnet}. Here, we simply leverage the pooling functionality to aggregate the concatenated tensor, that is to say, ``A'' in Figure \ref{fig:ortho1} is pooling in our current implementation. Actually, it can be designed to be other learnable modules to aggregate the tensor. We will further analysis on this in Section \ref{exp} and \ref{dis}.

\subsection{Training Objective}
Following DELG \cite{cao2020unifying}, the training of our method involves only one $L_2$-normalized $N$ class prediction head ${\cal{\hat{W}}}\in R^{512\times N}$ and just needs image-level labels. ArcFace margin loss \cite{deng2019arcface} is used to train the whole network: 
\begin{equation}\label{e3}
\mathit{L} = -\log\left ( \frac{\exp\left ( \gamma \times {AF} \left (  \hat{\omega}_{t}^{T}\hat{f_g},1\right )\right )}{\sum_{n}^{}\exp\left ( \gamma \times {AF} \left (  \hat{\omega}_{n}^{T}\hat{g},y_{n}\right )\right )} \right )
\end{equation}
where $\hat{\omega_i}$ refers to the $i_{th}$ row of $\cal{\hat{W}}$ and $\hat{f_g}$ is the $L_2$-normalized version of $f_g$. $y$ is the one-hot label vector and $t$ is the groundtruth class index ($y_t=1$). $\gamma$ is a scale factor. $AF$ denotes the ArcFace-adjusted cosine similarity and it can be calculated as $AF(s,c)$:
\begin{equation}\label{e2}
{AF}\left ( \mathit{s,c} \right ) = 
\left\{\begin{matrix}
\cos\left ( a\cos\left ( s \right ) +m \right ), & \mathit{if}\ c = 1\\ 
 s,&\mathit{if}\ c = 0 
\end{matrix}\right.
\end{equation}
where $s$ is the cosine similarity, $m$ is the ArcFace margin and $c=1$ means this is the groundtruth truth class.








\section{Experiments}
\label{exp}
\subsection{Implementation Details}
\textbf{Datasets and Evaluation metric}
Google landmarks dataset V2 (GLDv2) \cite{weyand2020google} is developed for large-scale and fine-grained landmark instance recognition and image retrieval. It contains a total of 5M images of 200K different instance tags. It is collected by Google to raise the challenges faced by the landmark identification system under real industrial scenarios as much as possible. 
Researchers from the Google Landmark Retrieval Competition 2019 further cleaned and revised the GLDv2 to be GLDv2-clean. It contains a total of 1,580,470 images and 81,313 classes. This dataset is used to train our models. To evaluate our model, we mainly use Oxford and Paris datasets with revisited annotations \cite{radenovic2018revisiting}, referred to be Roxf and Rpar in the following, respectively. There are 4,993 (6,322) images in the Roxf (Rpar) dataset and a different query set for each, both with 70 images. 
In order for a fair comparison with state-of-the-art methods \cite{noh2017large,cao2020unifying,ng2020solar}, mean average precision (mAP) is used as our evaluation metric on the Medium and Hard splits of both datasets.
mAP provides a robust measurement of retrieval quality across recall levels and has shown to have good discrimination and stability.

\textbf{Implementation details}
All the experiments in this paper are trained based on GLDv2-clean dataset. We randomly divide 80\% of the dataset for training and the rest 20\%  for validation. ResNet50 and ResNet101 are mainly used for experiments. Models are initialized from ImageNet pre-trained weights. The images first undergo augmentations by randomly cropping / distorting the aspect ratio; then, they are resized to $512 \times 512$ resolution. We use batch size of 128 to train our models on 8 V100 GPUs with 16G memory per card asynchronously for 100 epochs. One complete training phase takes about 3.8 days for ResNet50 and 6.3 days for ResNet101. SGD optimizer with momentum of 0.9 is used. Weight decay factor is set to 0.0001 and cosine learning rate decay strategy is adopted. Note that we train our models with 5 warming-up epochs and the initial learning rate is 0.05. For the ArcFace margin loss, we empirically set the margin $m$ as 0.15 and the ArcFace scale $\gamma$ as 30. For GeM pooling, we fix the parameter $p$ as 3.0.

As for feature extraction, following previous works \cite{noh2017large,cao2020unifying}, we use an image pyramid at inference time to produce multi-scale representations. Specifically, we use 5 scales, \ie, { 0.3535, 0.5, 0.7071, 1.0, 1.4142}, to extract final compact feature vectors. To fuse these multi-scale features, we firstly normalize them such that their $L_2$ norm equals 1, then the normalized features are averaged and finally a $L_2$ normalization is applied to produce the final descriptor. 

\subsection{Results}

\begin{table*}[!htbp]
\centering
\begin{tabular}{llcccccccc}
\toprule
\multicolumn{2}{c}{ \multirow{2}*{Method} }& \multicolumn{4}{c}{Medium} &\multicolumn{4}{c}{Hard}\\
\cline{3-6}  \cline{7-10} 
\multicolumn{2}{c}{} &Roxf &+1M & Rpar &+1M &Roxf &+1M & Rpar &+1M \\

\toprule
\multicolumn{10}{l}{\textsl{(A) Local feature aggregation + re-ranking} } \\ \hdashline
\multicolumn{2}{l}{HesAff-rSIFT-ASMK$\star$ +SP\cite{tolias2016image}}&60.6&46.80&61.40&42.30&36.70&26.90&35.00&16.80\\ 
\multicolumn{2}{l}{HesAff-HardNet-ASMK$\star$ +SP\cite{Mishkin2018Repeatability}}&65.60& - &65.20& - & 41.10 & - &38.50& - \\
\multicolumn{2}{l}{\textbf{HesAff–rSIFT–ASMK$\star$+SP$\rightarrow$R\cite{radenovic2018fine}–GeM+DFS\cite{dfs}}} &\textbf{79.10} &\textbf{74.30} &\textbf{91.00} &\textbf{85.90} &\textbf{52.70} &\textbf{48.70} &\textbf{81.00} &\textbf{73.20}\\ 
\multicolumn{2}{l}{DELF-ASMK$\star$ +SP\cite{noh2017large, radenovic2018revisiting}}&67.80&53.80&76.90&57.30&43.10&31.20&55.40&26.40\\ 	  			 		     
\multicolumn{2}{l}{DELF-R-ASMK$\star$ +SP\cite{teichmann2019detecttoretrieve}}&76.00&64.00&80.20&59.70&52.40&38.10&58.60&58.60\\ 	

\multicolumn{2}{l}{\textbf{R50-How-ASMK,n=2000\cite{tolias2020learning}}}&\textbf{79.40}& \textbf{65.80} &\textbf{81.60}& \textbf{61.80}  &\textbf{56.90} & \textbf{38.90} &\textbf{62.40}& \textbf{33.70} \\ 	
\toprule

\multicolumn{10}{l}{\textsl{(B) Global features} } \\ \hdashline
\multicolumn{2}{l}{R101-R-MAC\cite{gordo2017end} }&60.90  &39.30 & 78.90&54.80 &32.40 &12.50 &59.40 &28.00 \\ 
\multicolumn{2}{l}{R101-GeM $\uparrow$\cite{dsm}}&65.30  &46.10 &77.30 &52.60 &39.60 &22.20 &56.60 & 24.80\\ 
\multicolumn{2}{l}{R101-GeM-AP\cite{revaud2019learning}}&67.50  &47.50 &80.10 &52.50 &42.80 &23.20 &60.50 &25.10 \\ 
\multicolumn{2}{l}{R101-GeM-AP (GLDv1) \cite{revaud2019learning}}&66.30  & -  &80.20 & - &42.50 & - &60.80 &  - \\ 	
\multicolumn{2}{l}{R152-GeM\cite{radenovic2018fine}}&68.70  & - &79.70 & - & 44.20 & - & 60.30 & - \\ 
\multicolumn{2}{l}{ResNet101-GeM+SOLAR$\dagger$ \cite{ng2020solar}}&69.90  & 53.50 &81.60 &59.20 & 47.90 &29.90 & 64.50 & 33.40 \\ 
\multicolumn{2}{l}{R50-DELG\cite{cao2020unifying}}&69.70  &55.00 &81.60 &59.70 &45.10 &27.80 &63.40 &34.10 \\ 	
\multicolumn{2}{l}{\textbf{R50-DELG (GLDv2-clean)}\cite{cao2020unifying}}&\textbf{73.60}  &\textbf{60.60} &\textbf{85.70} &\textbf{68.60} &\textbf{51.00} &\textbf{32.70} &\textbf{71.50} &\textbf{44.40} \\ 

\multicolumn{2}{l}{\textbf{R50-DELG(GLDv2-clean){{$^r$}}}\cite{cao2020unifying}}&\textbf{77.51}  &\textbf{74.80} &\textbf{87.90} &\textbf{77.3} &\textbf{54.76} &\textbf{50.40}  &\textbf{73.82} &\textbf{61.01}  \\

\multicolumn{2}{l}{R101-DELG\cite{cao2020unifying}}&73.20  &54.80 & 82.40 &61.80 &51.20 &30.30 &64.70 &35.50 \\ 	
\multicolumn{2}{l}{\textbf{\textbf{R101-DELG(GLDv2-clean)}}\cite{cao2020unifying}}&\textbf{76.30}  &\textbf{63.70} &\textbf{86.60} &\textbf{70.60} &\textbf{55.60} &\textbf{37.50} &\textbf{72.40} &\textbf{46.90} \\ 
\toprule

\multicolumn{10}{l}{\textsl{(C) Global features + Local feature re-ranking} } \\ \hdashline
\multicolumn{2}{l}{R101-GeM$\uparrow$+DSM\cite{dsm}}&65.30  &47.60 &77.40 &52.80 &39.20 &23.20 &56.20 &25.00 \\ 	
\multicolumn{2}{l}{R50-DELG\cite{cao2020unifying}}&75.10  &61.10 &82.30 &60.50 &54.20 &36.80 &64.90 &34.80 \\ 
\multicolumn{2}{l}{\textbf{R50-DELG(GLDv2-clean)}\cite{cao2020unifying}}&\textbf{78.30}  &\textbf{67.20} &\textbf{85.70} &\textbf{69.60} &\textbf{57.90} &\textbf{43.60} &\textbf{71.00} &\textbf{45.70} \\ 

\multicolumn{2}{l}{\textbf{R50-DELG(GLDv2-clean){$^r$}}\cite{cao2020unifying}}&\textbf{79.08}  &\textbf{75.90}  &\textbf{88.78} &\textbf{77.69} &\textbf{58.40} &\textbf{52.40} &\textbf{76.20} &\textbf{61.60}  \\

\multicolumn{2}{l}{R101-DELG\cite{cao2020unifying}}&78.50  &62.70 &82.90 &62.60 &59.30 &39.30 &65.50 &37.00 \\ 
\multicolumn{2}{l}{\textbf{R101-DELG(GLDv2-clean)}\cite{cao2020unifying}}&\textbf{81.20}  &\textbf{69.10} &\textbf{87.20} &\textbf{71.50} &\textbf{\underline{64.00}} &\textbf{47.50} &\textbf{72.80} &\textbf{48.70} \\ 
\toprule

\multicolumn{2}{l}{\textbf{R50-DOLG (GLDv2-clean)}}&\textbf{\underline{80.50}}  & \textbf{\underline{76.58}} &\textbf{\underline{89.81}} & \textbf{\underline{80.79}}  &\textbf{\underline{58.82}} & \textbf{\underline{52.21}} &\textbf{\underline{77.7}} & \textbf{\underline{62.83}} \\ 	
\multicolumn{2}{l}{\textbf{R101-DOLG (GLDv2-clean)}}&\textbf{\underline{81.50}} & \textbf{\underline{77.43}} &\textbf{\underline{91.02}} &\textbf{\underline{83.29}} &\textbf{{61.10}} & \textbf{\underline{54.81}}  &\textbf{\underline{80.30}} &\textbf{\underline{66.69}}\\ 
\toprule
\end{tabular}
\caption{Results (\% mAP) of different solutions are obtained following the Medium and Hard evaluation protocols of Roxf and Rpar. ``$\star$'' means feature quantization is used and ``$\dagger$'' means second-order loss is added into SOLAR. ``GLDv1'', ``GLDv2'' and ``GLDv2-clean'' mark the difference in training dataset. $^r$ denotes our re-implementation. State-of-the-art performances are marked bold and ours are summarized in the bottom. The underlined numbers are the best performances.}
\label{t-1}
\end{table*}
\subsubsection{Comparison with State-of-the-art Methods}
We divide the previous state-of-the-art methods into three groups: (1) local feature aggregation and re-ranking; (2) global feature similarity search; (3) global feature search followed by re-ranking with local feature matching and spatial verification (SP). From some point of view, our method belongs to the global feature similarity search group. The results are summarized in Table \ref{t-1} and we can see that our solution consistently outperforms existing solutions.

\textbf{Comparison with local feature based solutions.} In the local feature aggregation group, besides DELF \cite{noh2017large}, it is worth mentioning that current work R50-How \cite{tolias2020learning} provides a manner for learning local descriptors with ASMK \cite{tolias2016image} and outperforms DELF. It achieves a boost up to 3.4\% on Roxf-Medium and 1.4\% on Rpar-Medium. 
However, the complexity of this work is considerable, where n=2000 shows it finally uses 2000 strongest local keypoints. 
Our method outperforms it by up 1.1$\%$ on Roxf-Medium and 8.21$\%$ on Rpar-Medium with the same ResNet50 backbone. 
For the hard samples, our R50-DOLG achieves 58.82\% and 77.7\% in mAP on the Roxf and Rpar respectively, which is significantly better than 56.9\% and 62.4\% achieved by R50-How. The results show that our single-stage model is better than existing local feature aggregation methods which are enhanced by a second re-ranking stage. 

\textbf{Comparison with global feature based solutions.}
Our method completes image retrieval with single-stage and the global feature based solutions do the same. It can be found the global feature learned by DELG \cite{cao2020unifying} performs the best. Especially when the models are trained using the GLDv2-clean dataset. Our models are also trained on this dataset and they are validated to be better than DELG. The performance is significantly improved by our solution. For example, with Res50 backbone, the mAP is 80.5\% v.s. 77.51\% on Roxf-Medium and 58.82\% v.s. 54.76\% on Rofx-Hard. Please note that, our R50-DOLG performs better than R101-DELG. These results well demonstrate the superiority of our framework.

\textbf{Comparison with global+local feature based solutions.} In the solutions where global feature is followed by a local feature re-ranking, R50/101-DELG is still the existing state-of-the-art method. Compared with the best result of DELG, our method R50-DOLG outperforms the R50-DELG with a boost of up to 1.42$\%$ on Roxf-Medium, 1.03$\%$ on Rpar-Medium, 0.42$\%$ on Roxf-Hard and 1.5$\%$ on Rpar-Hard. Our R101-DOLG outperforms R101-DELG with a boost of up to 0.3\% on Roxf-Medium, 3.82\% on Rpar-Medium and 7.5\% on Rpar-Hard. From these results, we can see, although 2-stage solutions can well promote their single stage counterparts, our solution combining both local and global information is a better choice. 

\textbf{Comparison in mP@10.}  We compare mP@10 in Table \ref{t-mP}. It shows the mP@10 performances of DOLG are better than 2-stage DELG$^r$ on both RPar and Roxf. Such results validate our single-stage solution is more precise than state-of-the-art 2-stage DELG, owing to the advantages of end-to-end training and free of error accumulation.  
\begin{table}[]
    \centering
    \begin{tabular}{c|c|c|c|c}
    \hline
         Model &Roxf-M &Roxf-H &Rpar-M &Rpar-H \\
         \hline
         R50-DELG$^r$ &90.79 &69.00 &95.57 &92.00 \\
         R50-DOLG &92.52 &71.14 &98.43 &93.71 \\
         \hline
    \end{tabular}
    \caption{Results of mP@10 of different methods.}
    \label{t-mP}
\end{table}

\textbf{``+1M'' distractors.} From Table \ref{t-1}, DOLG and 2-stage DELG$^r$ outperform the official 2-stage DELG by a large margin. This is reasonable. Firstly, the DELG$^r$ and our DOLG are both trained for 100 epochs while the official DELG is only trained for 25 epochs, so the original DELG features are not so robust, (w/o 1M distractors, DELG-Global$^r$ outperforms DELG-Global by 3.9 points in mAP on Roxf-M and Re-ranking on Rpar is even slightly worse than DELG-Global). When a huge amount of distractors exist, less robust global and local feature will result in severer error accumulation (DELG-Global$^r$ $>$ 2-stage DELG with ``+1M''). As a consequence, significant performance gap appears between our re-implemented DELG and its official version. From the last two rows, we see DOLG still outperforms 2-stage DELG$^r$ when +1M distractors exist. 

\begin{figure*}[t]
\centering
\includegraphics[width=0.8\textwidth]{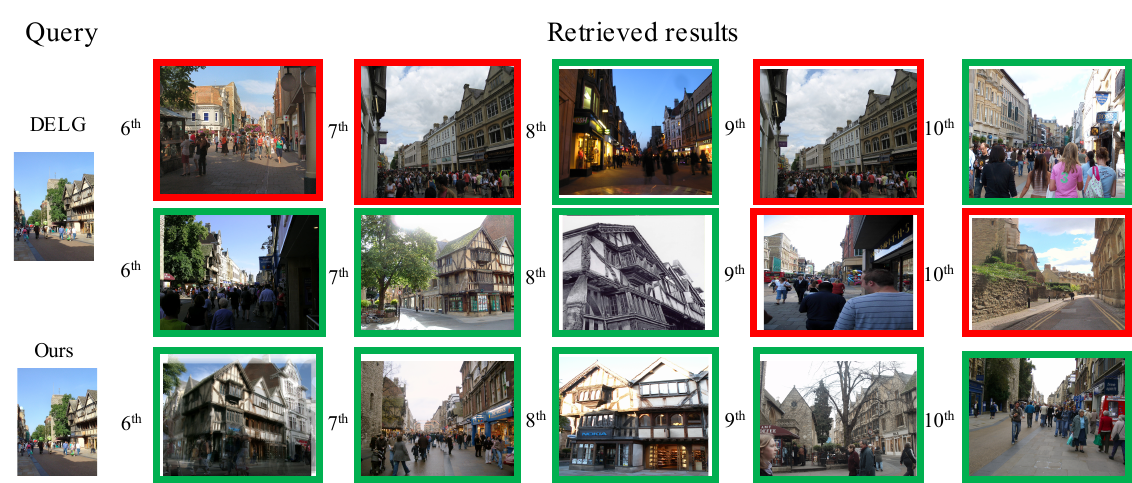}
\caption{Demonstration of top-10 retrieved results. The top-5 retrieved images are all correct and are excluded in this figure. Results of DELG global, DELG global+local and our DOLG are shown from top to bottom. Green and red boxes denote positive and negative images, respectively. }
\label{fig:demo}
\end{figure*}

\textbf{Qualitative Analysis.} We showcase top-10 retrieval results of a query image in Figure \ref{fig:demo}. We can see that state-of-the-art methods with global feature will result in many false positives which are semantically similar to the query. With re-ranking, some false positives can be eliminated but those with similar local patterns still exist. Our solution combines global and local information and is end-to-end optimized, so it is better at figuring out true positives.

\subsubsection{Ablation Studies}
To empirically verify some of our design choices, ablation experiments are conducted using the Res50 backbone.

\textbf{Where to Fuse.} To check which block is better for the global and local orthogonal integration, we provide empirical results to verify our choice. Specifically, shallow layers are known to be not appropriate for local feature representations \cite{noh2017large,cao2020unifying}, thus we mainly check the res3 and res4 block. 
We have implemented DOLG variants where the local branch(es) is (are) originated from $f_4$ only (both $f_3$ and $f_4$). Hence, fusing $f_3,f_4$ and $f_g$ means there are two orthogonal fusion branches based on Res3 and Res4, and the two orthogonal tensors generated from the two fusion branches are concatenated with $f_g$ and pooled. The results are summarized in Table \ref{t-2-location-ablation}. We can see that 1) without local branch, the global only setting performs worse. 2) Fusing $f_3$ or $f_4$ or  $both~f_3\&f4$ can improve the perform of ``Global only''. Fusing $f_3$ obviously outperforms fusing $f_4$ on Roxf although it is slightly worse on Rpar. Fusing both $f_3$ and $f_4$ does not provide improvement over $f_3$-$only$ but it is better than $f_4$-$only$. The above phenomena is reasonable. $f_3$ is of sufficient spatial resolution and its network depth is also sufficient, so it is better than $f_4$ to serve as local features. $both~f_3\&f_4$ will make the model more complicated. Besides, $f_g$ is derived from $f_4$ as well, then $both~f_3\&f_4$ setting may put more emphasis on $f_4$, therefore degrading the overall performance. Overall speaking, $f_3$-$only$ is the best. 

\textbf{Impact of Poolings.} In this experiment, we study how GeM pooling \cite{radenovic2018fine} and average pooling will make a difference to our overall framework. We report results of DOLG when the pooling function of the global branch and the orthogonal fusion module alters. With other settings kept the same, the performances of R50-DOLG are presented in Table \ref{t-5-pooling-ablation}. It is interesting to see that using GeM pooling for the global branch while using average pooling for the orthogonal fusion module results in the best combination. 

\begin{table}[t]
\centering
\begin{tabular}{llp{5mm}<{\centering}p{5mm}<{\centering}p{5mm}<{\centering}p{5mm}<{\centering}p{5mm}<{\centering}p{5mm}<{\centering}}
\hline
\multicolumn{2}{c}{ \multirow{2}*{Location} }& \multicolumn{3}{c}{Roxf} &\multicolumn{3}{c}{Rpar}\\
\cline{3-8} 
\multicolumn{2}{c}{} &E &M & H &E &M &H  \\
\toprule
\multicolumn{2}{l}{Global only}&90.65 &78.21 &56.31 &95.65 &89.00 &76.17  \\
\multicolumn{2}{l}{Fuse f4-only}&92.08 &79.39 &58.13 & 95.93&\textbf{89.92} & \textbf{77.92} \\ 
\multicolumn{2}{l}{Fuse f3-only}&\textbf{93.17} &\textbf{80.50} &\textbf{58.82} &95.95 &89.81 &77.70  \\ 
\multicolumn{2}{l}{both f3\&f4}&92.34 &79.41 &57.08 &\textbf{96.01} &89.78 &77.69  \\ 
\hline
\end{tabular}
\caption{Experimental results of DOLG variants where the orthogonal fusion is performed at different locations.}
\label{t-2-location-ablation}
\end{table}

\begin{table}[t]
\centering
\begin{tabular}{ll|p{5mm}<{\centering}p{5mm}<{\centering}p{5mm}<{\centering}p{5mm}<{\centering}p{5mm}<{\centering}p{5mm}<{\centering}}
\toprule
\multicolumn{2}{c|}{ Pooling }& \multicolumn{3}{c}{Roxf} &\multicolumn{3}{c}{Rpar}\\
\cline{1-8}
Global &Ortho &E &M & H &E &M &H \\
\hline
GeM &GeM &92.62 &78.28 &55.30 &96.20 &89.50 &76.99  \\ 
AVG &AVG &92.20 &78.14 &56.14 &95.86 &89.25 &76.32  \\ 
GeM &AVG &\textbf{93.17}	&\textbf{80.50}		&\textbf{58.82}	&\textbf{95.95}	&\textbf{89.81}	&\textbf{77.70}  \\ 
AVG &GeM &89.63  &73.48 &44.88 &94.67 &86.76 &72.98  \\ 
\hline
\end{tabular}
\caption{Differences when different pooling functions are used. ``AVG'' means ordinary global average pooling.}
\label{t-5-pooling-ablation}
\end{table}

\textbf{Impact of Each Component in the Local Branch.} A multi-atrous block and self-attention block are designed in our local branch to simulate the spatial feature pyramid by dilated convolution layers \cite{atrous} and to model the local feature importance with attention mechanism \cite{noh2017large}, respectively. We provide experimental results to validate the contribution of each of these components by removing individual component from the whole framework. The performance is shown in Table \ref{t-3-aspp+att-ablation}. It is clear that fusing the local features helps to improve the overall performance significantly. The mAP is improved from 78.2\% to 80.5\% and 89.0\% to 89.8\% on Roxf-Medium and Rpar-Medium, respectively. When Multi-Atrous module is removed, the performance will slightly drop on the Medium and Hard splits, especially for the hard split. For example, mAP is decreased from 58.82\% to 58.36\% and 77.7\% to 76.52\% on Roxf-Hard and Rpar-Hard, respectively. However, for easy cases, Multi-Atrous will make the performance slightly worse, but this make little difference because the mAP is already very high and the retrieval performance drop is very limited for easy case. Such results validate the effectiveness of Mutli-Atrous module. When the self-attention module is removed the performance also notably drops, which is consistent with results obtained by \cite{cao2020unifying}.
\begin{table}[b]
\centering
\begin{tabular}{llp{5mm}<{\centering}p{5mm}<{\centering}p{5mm}<{\centering}p{5mm}<{\centering}p{5mm}<{\centering}p{5mm}<{\centering}}
\toprule
\multicolumn{2}{c}{ \multirow{2}*{Config} }& \multicolumn{3}{c}{Roxf} &\multicolumn{3}{c}{Rpar}\\
\cline{3-5}  \cline{6-8} 
\multicolumn{2}{c}{} &E &M & H &E &M &H  \\
\toprule
\multicolumn{2}{l}{w/o Local}&90.65 &78.21 &56.31 &95.65  &89.00 &76.17  \\ 
\multicolumn{2}{l}{w/o MultiAtrous}&\textbf{93.48} &80.48 &58.36 & \textbf{96.66} &89.27 &76.52 \\ 
\multicolumn{2}{l}{w/o Self-ATT}&90.64 &78.15 &55.34 &95.73 &89.48 &77.16  \\ 
\multicolumn{2}{l}{Full Model}&93.17 &\textbf{80.50} &\textbf{58.82} &95.95 &\textbf{89.81} &\textbf{77.70}\\
\hline
\end{tabular}
\caption{Ablation experiments on components of the local branch in our framework. }
\label{t-3-aspp+att-ablation}
\end{table}

\textbf{Verification of the Orthogonal Fusion.} In the orthogonal fusion module, we propose to decompose the local features into two components, one is parallel to the global feature $f_g$ and the other is orthogonal to $f_g$. Then we fuse the complementary orthogonal component and $f_g$. To show such orthogonal fusion is a better choice, we conduct experiments by removing the orthogonal decomposition procedure shown in Figure \ref{fig:ortho1} and concatenate the $f_l$ and $f_g$ directly. We also try fusing $f_l$ and $f_g$ by Hadamard product (also known as element-wise product), which is usually used to fuse two vectors. We can find from the empirical results (see Table \ref{t-4-orth-ablation}) that among the three fusion schemes, our proposed orthogonal fusion performs the best. Such experimental results are also within our expectation. With orthogonal fusion, the information relevant to the global feature $f_g$ is excluded from each local feature point $f_l^{(i,j)}$. In this way, the output local feature points are the most informative one and are orthogonal to $f_g$. Not only will they provide complementary information to better describe a image, but also they will not put extra emphasis on global feature $f_g$ because of their irrelevance. 

\begin{table}[t]
\centering
\begin{tabular}{llp{5mm}<{\centering}p{5mm}<{\centering}p{5mm}<{\centering}p{5mm}<{\centering}p{5mm}<{\centering}p{5mm}<{\centering}}
\toprule
\multicolumn{2}{c}{ \multirow{2}*{Method} }& \multicolumn{3}{c}{Roxf} &\multicolumn{3}{c}{Rpar}\\
\cline{3-5}  \cline{6-8} 
\multicolumn{2}{c}{} &E &M & H &E &M &H  \\
\toprule
\multicolumn{2}{l}{Concatenation$^\dagger$}& 91.29 & 78.40 & 56.55 & 95.88 &89.37  &76.80 \\ 
\multicolumn{2}{l}{Hadamard}&92.21 &79.20 &56.76 & 95.94 &\textbf{89.91} & 77.40 \\ 
\multicolumn{2}{l}{orthogonal}&\textbf{93.17} &\textbf{80.50} &\textbf{58.82} &\textbf{95.95} &{89.81} &\textbf{77.70} \\ 
\hline
\end{tabular}
\caption{Comparison of orthogonal fusion with other fusion strategies. Concatenation and Hadamard product are explored. $^\dagger$ with $m=2.0,\gamma=30$ for the ArcFace margin loss,  otherwise the training does not converge. }
\label{t-4-orth-ablation}
\end{table}

\section{Discussions}
\label{dis}
Here, we would like to have some discussion on our \textbf{current implementations} and \textbf{model complexity}. First of all, we have not extensively studied and tuned on many hyper-parameters, such as $p$ for GeM, $\gamma$ and $m$ for ArcFace margin loss, and the dilation rate $s$ settings of dilated convolution layers. Instead, we directly follow the practices in DELG \cite{cao2020unifying} and ASPP \cite{atrous}. We do so in order to show the effectiveness of our proposed building blocks instead of tuning for better models, although we believe tuning on these parameters may obtain better performance. The other thing worthy of mention is the orthogonal fusion module. We pay our attention to developing single stage solution by aggregating orthogonal local and global information. The design choice of the aggregation operation denoted as ``A'' in Figure \ref{fig:ortho1} is simply chosen from GeM and average pooling for proof-of-concept purpose. Note that average pooling is a linear operation, in this case, the orthogonal fusion module is equivalent to pool the local feature at first and then perform the projection and subtraction, so its computation can be further simplified. In short, our current orthogonal fusion module is sufficiently simple yet effective. We believe exploring more complicated learning based aggregation ``A'' in Figure \ref{fig:ortho1} is promising and it is left as our future work. As for the complexity, compared to DELG \cite{cao2020unifying} and DELF \cite{noh2017large}, extra computational cost comes from the Multi-Atrous module and the orthogonal fusion module. The former one is composed of a few dilated convolution layers meanwhile the latter one can currently be reduced to $Pool(f_l) - (Pool(f_l)\cdot f_g)f_g/|f_g|^2$. Therefore, the overhead of our solution is quite limited. Besides, our retrieval process can be finished in a single-stage.  

\section{Conclusion}
In this paper, we make the first attempt to fuse local and global features in an orthogonal manner for effective single-stage image retrieval. We have designed a novel local feature learning branch, where multi-atrous module is leveraged to simulate spatial feature pyramid to handle scale variation among images and self-attention module is adopted to perform significance modeling for each local descriptor. We also design a novel orthogonal fusion module in order to combine complementary local and global information, to mutually reinforce each other and produce a representative final descriptor via objective-oriented training. Extensive experimental results have been shown for proof-of-concept purpose, and we also significantly improve state-of-the-art performance on Roxf and Rpar.

\balance
{\small
\bibliographystyle{ieee_fullname}
\bibliography{egbib}
}

\end{document}